\pdfoutput=1

\documentclass[11pt]{article}

\usepackage[preprint]{acl}

\usepackage{times}
\usepackage{latexsym}

\usepackage[T1]{fontenc}
\usepackage{balance}
\usepackage[utf8]{inputenc}
\usepackage{hyperref}
\usepackage{microtype}
\usepackage{balance}

\usepackage{inconsolata}

\usepackage{graphicx}

\usepackage{amsmath}
\usepackage{amsthm}
\usepackage{booktabs}
\usepackage{algorithm}
\usepackage{algorithmic}
\usepackage{tikz}
\usetikzlibrary{shadows} 
\usepackage{tcolorbox} 
\usepackage{fancyvrb}
\usepackage{listings}
\usepackage{pifont} 
\usepackage{spverbatim}
\usepackage{enumitem}
\usepackage{setspace}
\usepackage{multirow}

\usepackage[table,xcdraw]{xcolor}  

\usepackage{enumitem}
\setlist{leftmargin=*, nosep} 
\usepackage{placeins} 

\usepackage[english, status=draft, margin=false, inline=true]{fixme}
\fxusetheme{color}

\definecolor{forestgreen}{RGB}{34,139,34} 

\newcommand{\rmspace}{\vspace{-2ex}}

\newcommand*\circled[1]{\tikz[baseline=(char.base)]{
            \node[shape=circle,draw,inner sep=.6pt] (char) {#1};}}

\title{TextMineX: Data, Evaluation Framework and Ontology-guided LLM Pipeline for Humanitarian Mine Action}

\author{
Chenyue Zhou$^{1,2,3}$ \quad Gürkan Solmaz$^{1}$ \quad Flavio Cirillo$^{1}$ \quad Kiril Gashteovski$^{1,5}$ \quad Jonathan Fürst$^{4}$ \\
$^{1}$NEC Laboratories Europe, Germany
$^{2}$University of Stuttgart, Germany\\ 
$^{3}$VAGO Solutions, Germany 
$^{4}$Zurich University of Applied Sciences, Switzerland \\
$^{5}$CAIR, Ss. Cyril and Methodius University of Skopje, North Macedonia \\
\texttt{zhou@vago-solutions.ai}, \texttt{jonathan.fuerst@zhaw.ch}\\
\texttt{guerkan.solmaz,flavio.cirillo,kiril.gashteovski@neclab.eu}\\
}

\begin{document}
{\makeatletter\acl@finalcopytrue
  \maketitle
}

\begin{abstract}
Humanitarian Mine Action (HMA) addresses the challenge of detecting and removing landmines from conflict regions. Much of the life-saving operational knowledge produced by HMA agencies is buried in unstructured reports, limiting the transferability of information between agencies. To address this issue, we propose TextMineX: the first dataset, evaluation framework and ontology-guided large language model (LLM) pipeline for knowledge extraction from text in the HMA domain. TextMineX structures HMA reports into (subject, relation, object)-triples, thus creating domain-specific knowledge. To ensure real-world relevance, we utilized the dataset from our collaborator Cambodian Mine Action Centre (CMAC). We further introduce a bias-aware evaluation framework that combines human-annotated triples with an LLM-as-Judge protocol to mitigate position bias in reference-free scoring. Our experiments show that ontology-aligned prompts improve extraction accuracy by up to 44.2\%, reduce hallucinations by 22.5\%, and enhance format adherence by 20.9\% compared to baseline models. We publicly release the dataset and code\footnote{\url{https://github.com/nec-research/TextMineX}}.
\end{abstract}

\section{Introduction}
\label{sec:introduction}
Humanitarian Mine Action (HMA)---detecting and removing landmines from past (and ongoing) conflicts in order to return land to civilian use---remains a critical humanitarian challenge: in 2022 alone, there were 4,710 casualties globally, 85\% of which were civilians \citep{UNMineAwareness2025,landmineMonitor2023}. Over the decades, HMA authorities have published large amount of life-saving knowledge, yet much of it remains locked away in unstructured, free-form text reports, thus making this knowledge largely inaccessible. The automatic extraction and organization of this important information is, therefore, not only a technical advancement, but also a humanitarian imperative that can make life-saving insights more accessible, actionable and transferable across many humanitarian agencies.

\begin{figure}
    \centering
    \includegraphics[width=0.45\textwidth]{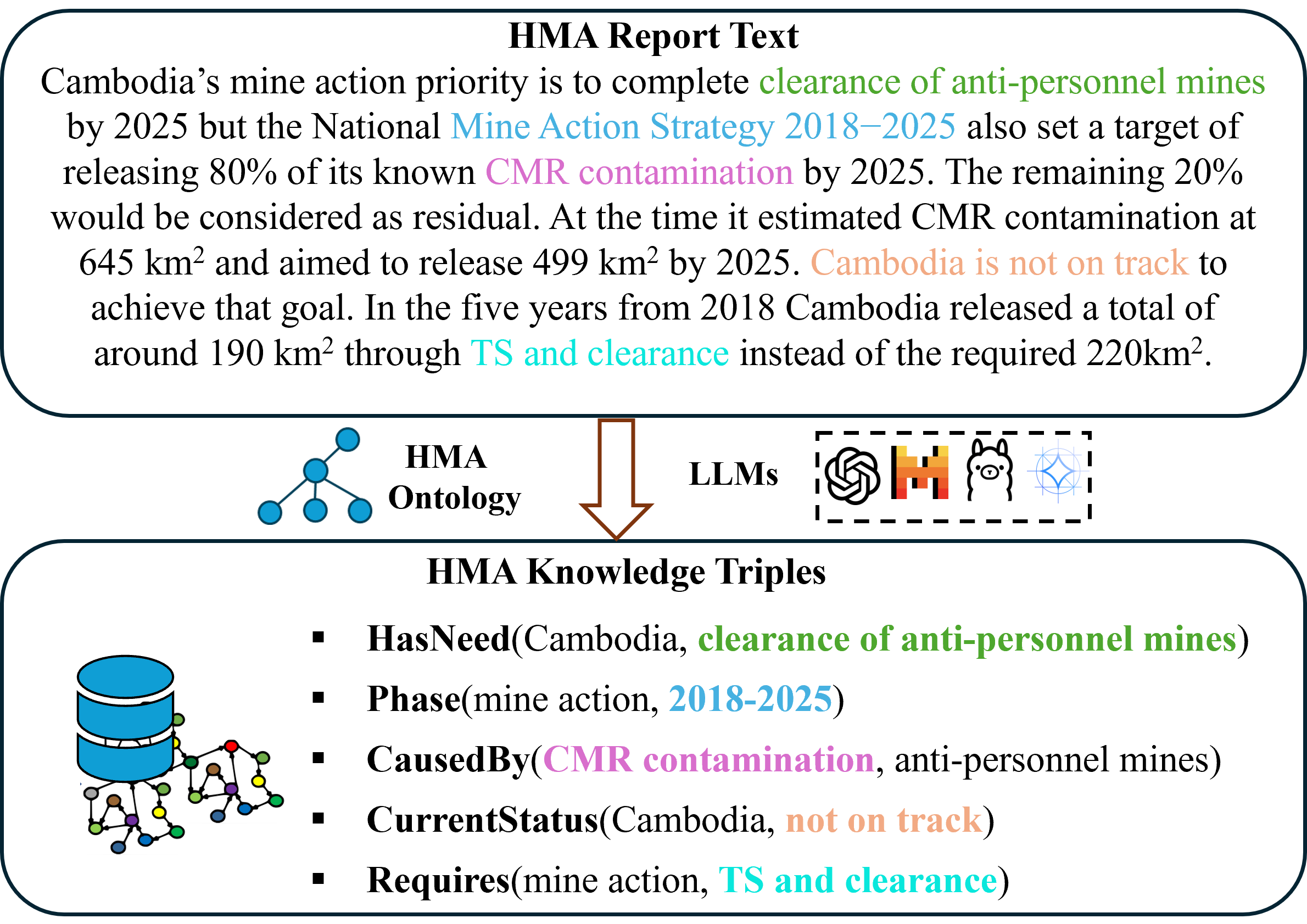}
    \caption{Example knowledge graph triple extraction from texts, guided by the HMA ontology. The task is to extract as many triples as possible while ensuring ontology conformance and source-text faithfulness.}
    \label{fig:triple}
    \rmspace
\end{figure}

To make HMA knowledge accessible, we turn to Information Extraction (IE). Prior research has shown that structuring information from natural language text data is useful for many downstream tasks, such as question answering \cite{xiong2024interactive}, fact retrieval \cite{han2023improving} or predicting new facts that were not present originally in the data \cite{broscheit2020can}. More importantly, IE pipelines are useful frameworks for quick information exchange between organizations \cite{poletto2021developing}, which is of vital importance for HMA, because the work is typically done in various agencies from different countries.

To effectively structure such domain-specific information, researchers have proposed data, benchmarks and methods for IE in various domains, including finance \cite{hamad2024fire}, material science \cite{cheung2024polyie}, medicine \cite{romero2025insightbuddy} and humanitarian crisis response \cite{fekih2022humset}. Although LLMs have become ubiquitous, it has been shown that using off-the-shelf LLMs for domain-specific IE is not an optimal approach \cite{pang2024uncovering,farzi2024get,dagdelen2024structured}. Therefore, there is a need for domain-specific IE applications.

In this work, we propose TextMineX: the first dataset, evaluation framework and pipeline for IE for the HMA domain. The task is to extract information from HMA reports in the form of (subject, relation, object)-triples, thus creating structured knowledge for the HMA domain (Fig.~\ref{fig:triple}). The application provides demining technical information in the form of knowledge triples that can be stored, shared, and queried across demining agencies. Prior data and benchmarks \citep[\emph{inter alia}]{mihindukulasooriya2023text2kgbench} rely on toy data schemas (i.e., ontologies), which limits their applicability to the complex and specialized HMA documents. By contrast, we utilized the publicly available technical reports from our collaborator Cambodian Mine Action Centre (CMAC). CMAC has been initiated by the UN more than 30 years ago, is one of the world's most experienced and effective demining organizations. Its collaborations within the UN and the Geneva International Centre for Humanitarian Demining (GICHD), ensure that the data is useful in real-world HMA applications beyond Cambodia~\cite{MonitorCambodia2023, UNDPCambodia2021}. While there is evidence that ontology-guided methods are effective \citep{cauter2024ontology}, they are limited to single-sentence inputs and overlook context-level reasoning. TextMineX addresses this gap by enabling context-level reasoning, while aggregating a set of HMA-related ontologies. Finally, no framework is bias-aware, which might distort the evaluation results, a gap that TextMineX also addresses.

To sum up, our contributions are: \circled{1} \textbf{Data:} we introduce a curated dataset, curated ontology and annotated ground-truth data for humanitarian demining operations, systematically categorizing operational entities and relationships. \circled{2} \textbf{Evaluation:}  We evaluate extracted triples against our annotated dataset and introduce a bias-aware LLM-as-Judge framework for reference-free scoring. Experiments on closed and open LLMs show that position bias skews rankings. 

\circled{3} \textbf{Knowledge Extraction:} We propose a prompt-based pipeline that combines layout-aware document chunking, ontology-guided extraction, and multi-perspective evaluation. To our knowledge, this is the first LLM application of knowledge extraction for HMA. \circled{4} \textbf{In-Context Learning Optimization:} We found that prompts enriched with ontology-aligned examples improve triple extraction accuracy by up to 44.2\%, reduce hallucinations by 22.5\%, and enhance format conformance by up to 20.9\% compared to baseline LLMs. These findings provide practical insights for prompt construction.

\section{Related Work}
\label{sec:Related Work}
LLMs, such as those from the GPT family \cite{hurst2024gpt}, Llama \citep{touvron2023llama}, BLOOM \citep{workshop2022bloom}, and PaLM \citep{chowdhery2023palm}, have  transformed the field of knowledge extraction from text. They possess advanced language understanding and reasoning capabilities, making them well-suited for extracting knowledge from unstructured text or well-structured documents~\cite{colakoglu2025problemsolvedinformationextraction}, especially when paired with prompting techniques like in-context learning (ICL) \citep{brown2020language}.

ICL enables LLMs to learn new tasks by providing input-output demonstrations during inference. Depending on the number of examples provided, this can range from zero-shot (no demonstrations) to one-shot or few-shot learning (multiple demonstrations) \citep{min2022rethinking, liu2023pre}. This method enhances the models' ability to generalize from minimal data. \citet{zhu2023llms} showed the effectiveness of ICL for knowledge extraction. \citet{mihindukulasooriya2023text2kgbench} introduced an approach that utilizes ontology guidance to extract knowledge from text. Their work highlights the potential of LLMs in extracting domain-specific knowledge constrained by ontological rules. In our study, we adapt this approach to the domain of humanitarian demining, employing a set of specialized ontologies to guide the extraction process.

Evaluating generated texts is a challenging task, especially when limited ground-truth data are available. To address this problem, recent approaches include multi-faceted fact-based evaluation~\cite{gashteovski-etal-2022-benchie}, generating synthetic data to train an evaluator model \citep{saad2023ares,kim2025evaluating}, annotating datasets using a human-in-the-loop methodology \citep{dagdelen2024structured}, or leveraging strong LLMs as judges \citep{zheng2024judging, bavaresco2024llms}. Our work involves annotating extracted triples to create an evaluation dataset, applying LLMs as judges and analyzing the alignment between these two evaluation methods.
For more detailed discussion on related work, see Appendix \ref{app:rel-work}.

\section{Benchmark Dataset Creation}
\label{sec:Annotation}

\begin{figure}
    \centering
    \includegraphics[width=1.0\linewidth]{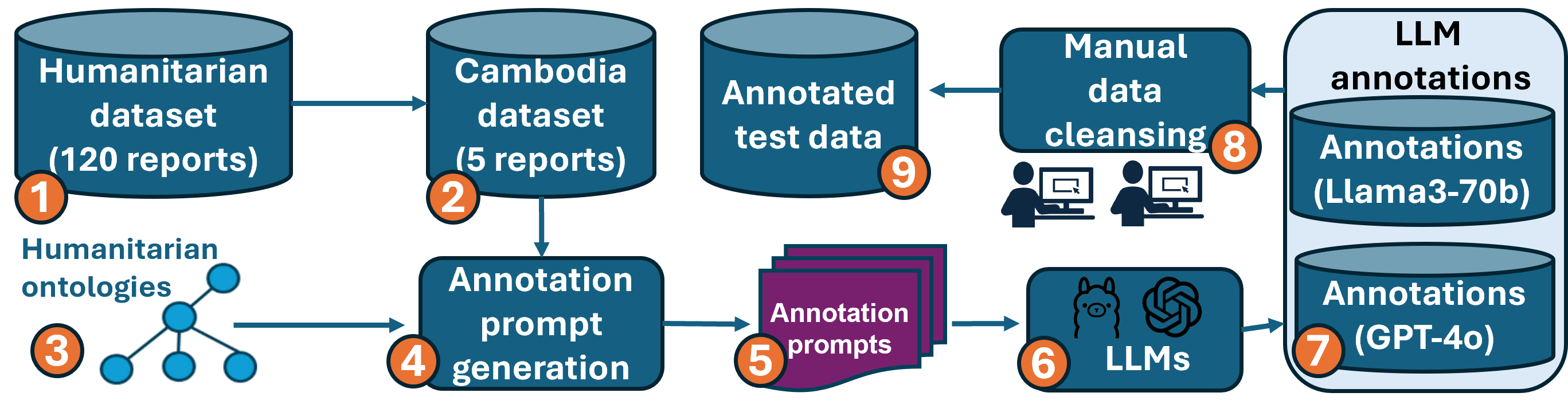}
    \caption{Semi-automatic creation of the humanitarian mine action dataset. The dataset contains a large and diverse set of technical reports from the global mine action and a smaller LLM- and human-annotated portion for Cambodian mine action.} 
    \label{fig:annotation}
    \rmspace
\end{figure}

We illustrate the process of creating the annotated humanitarian dataset in Fig.~\ref{fig:annotation}.
We curated 120 online available technical reports: 60 reports from the Cambodian Mine Action Center (CMAC)\footnote{https://cmac.gov.kh/publications/} and 60 from Geneva International Centre for Humanitarian Demining (GICHD) websites\footnote{https://www.gichd.org/publications-resources/publications/} \circled{1}. The reports are all in the PDF format. The curation is based on relevance to the humanitarian demining, language (English), and recency. From the initial humanitarian dataset of 120 PDF documents \circled{1}, we filtered a dataset of five  m most recent mine action reports from CMAC of overall 233 pages (see Table~\ref{tab:dataset_stats}) \circled{2}. 

\begin{table}
\centering
\scriptsize
\caption{Dataset Statistics for TextMineX Corpus}
\label{tab:dataset_stats}
  \resizebox{\linewidth}{!}{%
\begin{tabular}{lrrrrr}
\toprule
\textbf{Document} & \textbf{Pages} & \textbf{Chars} & \textbf{Words} & \textbf{Sent.} & \textbf{Nums} \\
\midrule

Annual progress report & 170 & 347,249 & 49,164 & 1,887 & 3,672 \\

Mine clearance report & 14 & 57,537 & 8,908 & 452 & 1,222 \\
Integrated work plan & 21 & 44,952 & 6,153 & 323 & 611 \\
Cluster munition remnant report & 9 & 37,846 & 5,865 & 327 & 893 \\
Article 7 report & 19 & 35,139 & 5,003 & 164 & 920 \\
\midrule
\textbf{Total} & 233 & 522,723 & 75,093 & 3,153 & 7,318 \\
\bottomrule
\end{tabular}}
\label{tab:dataset_stats}
\rmspace
\end{table}

Simultaneously, we selected domain-specific ontologies through working with domain experts, by first incorporating data models from the Information Management System for Mine Action (IMSMA)\circled{3}.
We performed a survey of demining related ontologies together with domain experts and we incorporate six ontologies from Information Management System for Mine Action (IMSMA) Core\footnote{\url{https://www.gichd.org/our-response/information-management/imsma-core/}}.
These ontologies are used for demining information system but not strictly for HMA. Furthermore, we add a more general humanitarian domain ontology from \textit{Empathi} \citep{8665539} to make the overall HMA ontology more comprehensive. We filter out the concepts that are not relevant to HMA from Empathi by utilizing knowledge of subject-matter expert. For instance, certain concepts related to natural disasters (e.g., flood response) are not directly related to demining. As a result, HMA ontology integrates seven ontologies (160 entity types, 86 relation types) covering diverse aspects of HMA (see details in Sec.~\ref{app:ontology}).

The annotation prompt generation \circled{4} first parses reports into text chunks, then systematically combines each chunk with ontology templates. Each template specifies the relation types for a specific domain (e.g., mine action events, land contamination, mine clearance). This generates 2,520 prompts (360 chunks $\times$ 7 templates), where each prompt contains both report context and ontology specifications for that domain.

We randomly select 100 prompts \circled{5} out of the 2,520, and apply them with the same prompts using GPT-4o and Llama3-70B \circled{6} to initially annotate the data as the ``LLM annotations\circled{7}. 

For each prompt and data chunk pair, each LLM generates a set of knowledge triples as outputs. Then, a human annotator who is an expert in natural language processing working with knowledge graphs reviewed these outputs, filtering out incorrect knowledge triples and aggregating the valid ones in a new set. Incorrect knowledge triples can include triples following wrong format, mixed order of entities, hallucinated entities/relations (e.g., triples that are not included in the data chunk) and so on. In addition, there exist duplicate triples which are removed to create the final reference set. Those included 1,095 unique triples across the 100 prompts\circled{8}. The end result of the human annotation is a set of knowledge triples which are relevant and correct without hallucination or formatting problems, faithful to the given data chunk. This clean set comprises the ``annotated test data'' \circled{9}. The annotated test data can serve as the ground-truth for evaluations of performance of different techniques and LLMs.

\label{sec:Annotation}

Finally, to assess the reliability of the annotations, a subject-matter expert---working in the demining innovation area---annotated 20 sets of data (corresponding to 286 knowledge triples) following the same methodology. Because we are interested in the overlap between the triples extracted by two independent annotators, we quantify the inter-annotator agreement with Jaccard Index, Dice Coefficient and Overlap Coefficient for these 20 sets. The two annotators have on average 0.89, 0.94, and 0.97 agreements respectively. These consistently high agreement scores indicate that the annotations in the full dataset are of similarly high quality.

\section{TextMineX Overview}
\label{sec:Methodology}

\subsection{Humanitarian Mine Action Task}
Knowledge triple extraction from HMA reports can be formally defined as follows: given an ontology $\mathcal{O} = (\mathcal{E}, \mathcal{R})$, where $\mathcal{E}$ is a set of entities and $\mathcal{R}$ is a set of relations, and a textual context $C$, the objective is to design a prompt $P(C, \mathcal{O})$ that guides an extractor model $M$ to extract triples $T = \{(s_1, r_1, o_1), \cdots, (s_n, r_n, o_n) \mid s, o \in \mathcal{E},\, r \in \mathcal{R}\}$, where $n$ is the number of extracted triples, and subjects/objects ($s_i / o_i$), and relations ($r_i$) are extracted from the source text and mapped to $\mathcal{E}$ and $\mathcal{R}$ respectively. 
The extracted triples must remain consistent with both the ontology and the source text; i.e.~ $T = M\bigl(P(C, \mathcal{O})\bigr)$.
\begin{figure*}
    \centering
    \includegraphics[width=\textwidth]{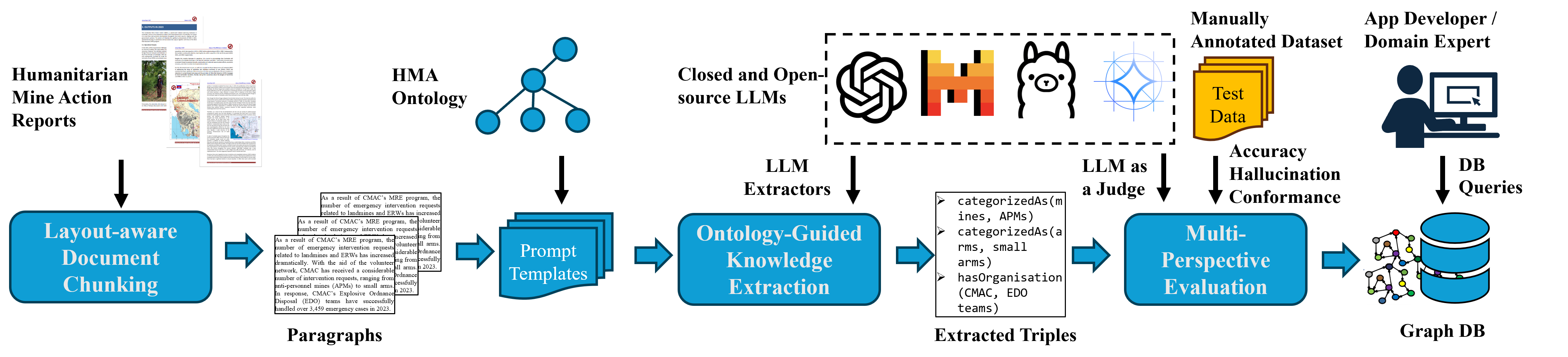}
\caption{\textbf{TextMineX Overview.} 
Reports are preprocessed into paragraph chunks, used as test inputs during inference. Each chunk is combined with an instruction template and ontology, then passed through LLMs for triple extraction. We apply a multi-perspective evaluation using both reference-based and reference-free methods. Extracted triples are stored in a database, queried by developers and HMA domain experts.}
    \label{fig:pipeline}
    \rmspace
\end{figure*}

\subsection{Overall Pipeline}

Figure~\ref{fig:pipeline} shows our triple extraction method. In the \textbf{Layout-Aware Document Chunking} phase, PDF reports are split into paragraph chunks. These are used with a newly constructed HMA ontology in the \textbf{Ontology-Guided Knowledge Extraction} phase. For evaluation, we apply a \textbf{Multi-Perspective Evaluation} combining reference-based metrics on our annotated dataset and a reference-free LLM-as-a-Judge approach. All LLM calls use greedy decoding (temperature = 0, top\_p = 1.0) to ensure deterministic outputs.

\paragraph{Layout-Aware Document Chunking} \label{preprocess}
The input to our pipeline is PDF-formatted demining reports, which contain rich human-readable structures (chapters, sections, tables, lists, and figures). To prepare them for LLM consumption, we segment each document into semantically coherent chunks that preserve context while fitting within typical model context windows \citep{liu2024lost}. We leverage Open-Parse’s document understanding capabilities \citep{smock2021tabletransformer} to identify layout elements and extract paragraph-level segments. On our reports, this yields chunks averaging 127 words (std.\ 6), which aligns well with both small and large LLM context limits.

\paragraph{Ontology-Guided Knowledge Extraction}
Given the text chunks as input, our goal is to extract $(s, r, o)$-triples. First, the extracted triples must be accurate w.r.t. the source text. An ideal extraction system would extract all triples (recall=1) precisely (precision=1). Second, extracted triples must conform to a specified ontology so that extracted demining knowledge can be stored and shared between organizations in a compatible way. We address this by combining a domain-specific HMA ontology with LLM in-context learning to extract triples from paragraphs.

\paragraph{Prompt Templates}
We design five prompting strategies for knowledge triple extraction:
(1) \emph{Zero-shot} (instruction and context only), 
(2) \emph{One-shot with Random Sentence (RS)}, 
(3) \emph{One-shot with Random Paragraph (RP)}, 
(4) \emph{One-shot with Ontology-Aligned Sentence (OS)}, and 
(5) \emph{One-shot with Ontology-Aligned Paragraph (OP)}. 
One example prompt is provided in Appendix.

\emph{Ontology-aligned} demonstrations share the same ontology (entity types and relations) as the target context, while \emph{random} demonstrations use unrelated ontology. Sentences are extracted from paragraphs using NLTK sentence tokenizer\footnote{\url{https://www.nltk.org/howto/stem.html}}. For OS and OP prompts, we design a retrieval algorithm that selects demonstrations from our annotated dataset (\S~\ref{sec:Annotation}) by identifying the shortest context-answer pair matching the target ontology, for minimizing token costs while retaining high semantic similarity. To prevent data leakage and ensure a fair evaluation, we implement a second retrieval step if the initially-retrieved demonstration contains the same or part of target context as the test instance. In such cases, we select the next shortest matching example instead. This ensures that the retrieved demonstrations do not overlap with the evaluation context, preserving the integrity of the inference.

We design these prompt templates to test two hypotheses:
(a) \textit{Ontology alignment enhances accuracy} by priming the model with ontology-specific reasoning. Semantically aligned demonstrations help constrain the label space and improve precision \citep{min2022rethinking,longdoes}. This is evident in our results, as the contrast between RS/RP and OS/OP confirms the benefit of ontology alignment. (b) \textit{Paragraph-level context improves extraction performance} by providing richer demonstrations that reflect how entities and relations are introduced across sentences in real-world reports. However, our results do not support this hypothesis as comparisons between RS vs.~RP and OS vs.~OP show no consistent improvement from paragraph-level context.

\subsection{Multi-Perspective Evaluation}

\paragraph{Reference-based Evaluation} 
HMA is a high-stakes decision-making domain, where incorrect triples---especially hallucinated ones---can misinform demining operations, leading to inefficiency. A comprehensive evaluation of triple extraction requires assessing accuracy, reliability, and structural validity. We evaluate models across three dimensions: (1) Triple Extraction Accuracy, (2) Hallucination Rate, and (3) Format Conformance. 

\paragraph{Triple Extraction Accuracy} Triple extraction accuracy serves as the primary metric, as it directly measures how well models extract knowledge triples from text. However, accuracy alone does not fully capture model reliability. The hallucination rate evaluates faithfulness by detecting extraneous or fabricated information, while format conformance ensures that outputs adhere to a syntactically valid structure, enabling seamless integration into downstream applications. We employ N-gram matching-based metrics such as BLEU \citep{papineni2002bleu}, ROUGE \citep{lin2004rouge}, and METEOR \citep{banerjee2005meteor} to assess the extracted triples. Additionally, we incorporate BERTScore \citep{zhang2019bertscore}, which leverages word embeddings to capture semantic similarity beyond lexical overlap.
To enhance the accuracy of our metrics, we first apply stemming and lemmatization by NLTK to normalize morphological variations. We then compute the accuracy metrics against the manually annotated test set.

\paragraph{Hallucination Rate}  
Accuracy measures how well the extracted triples match reference triples, but it does not fully capture whether the generated content is grounded in the input. A model could produce plausible triples that are semantically similar to the original text---thus achieving a high accuracy score---yet still be incorrect; i.e.~they do not appear in the input but seem reasonable. Hallucination is a prevalent issue~\citep{ji2023survey, huang2023survey, xu2024hallucination} and a critical aspect of our evaluation. To quantify this, for each extracted triple \(t = (s, r, o)\) we first normalize \(s\), \(r\), and \(o\), as well as the entire input context (tokenization, lemmatization, lowercasing, and punctuation removal). 
We flag $t$ as a hallucination if the normalized subject $s$ and object $o$ are not found as contiguous substrings in the normalized report text, or if the normalized relation $r$ is absent from the ontology set.
This procedure ensures that even “plausible” but unsupported triples are detected and penalized, thereby maintaining faithfulness and trustworthiness.

\paragraph{Format Conformance} Format conformance metric assesses whether the generated triples adhere to the correct syntactic format of \( r(s, o) \), where \( r \) is the relation and \( s \) and \( o \) denote the subject and object, respectively. We consider a triple well-formatted if it follows this structure. We accommodate edge cases where the subject or object contains numerical values with commas, such as {\small \textit{hasReliabilityInfo(2,500,011 square meters, landmine/ERW affected areas)}}, or phrases in parentheses, such as {\small \textit{hasAccidentOrganisationInfo(Quality of Life Survey (QLS), Department of Victim Assistance of CMAA)}}. Format conformance ensures that extracted triples follow a structured format necessary for practical use. A model with high accuracy but poor format conformance may fail to produce usable outputs, limiting its applicability in real-world.

\paragraph{Combined Score} To unify evaluation metrics into a single representative score, we apply min-max normalization and compute the overall \textit{Combined Score} as:
\begin{align}
S_{\text{combined}} = \frac{1}{k} \big(& S_{\text{BLEU}}' + S_{\text{ROUGE}}' + S_{\text{METEOR}}' \notag \\
& + S_{\text{BERTScore}}' + (1 - S_{\text{Hallucination}}') \big)
\end{align} where \( S' \) represents the normalized metric values within \([0,1]\), and the hallucination rate is inverted to penalize more hallucinations. $k=5$ is the number of metrics included in the score. Format conformance is excluded in the Combined Score, as our experimental results show consistently high format conformance across all extraction models and prompts, making it non-differentiating. \textit{Combined Score} provides a holistic measure of extraction quality while mitigating scale differences among individual metrics.

\paragraph{Reference-Free Evaluation}  
Evaluating generated texts is particularly challenging in domains like HMA, where annotated datasets are scarce. Demining reports are highly technical and domain-specific, requiring extracted triples to align with predefined ontologies of landmine types, clearance operations, and affected areas. Constructing a manually labeled test set is time-consuming and resource-intensive, limiting large-scale reference-based evaluation. Given these constraints, we explore an LLM-as-a-Judge approach as a potential reference-free evaluation framework for evaluating extracted triples. LLM-as-a-Judge offers a potential alternative to evaluation when ground-truth data is limited \citep{friel2023chainpoll,saad2023ares,es2023ragas}. The ultimate objective of our approach is to find an optimal judge LLM setting where the LLM consistently identifies the best candidate answer and provides a reasoned justification for its decision. 

We try to find the optimal LLM Judge setting by conducting systematic ranking experiments and analyzing correlations between the LLMs judged ranking and reference-based rankings. For these ranking experiments, we design Judge Prompts that instruct LLMs on evaluation criteria. We use five models: Mistral-7B \cite{jiang2023mistral7b}, Llama3-8B \cite{grattafiori2024llama}, Gemma2-9B \cite{team2024gemma}, LLaMA3-70B and GPT-4o \cite{hurst2024gpt}, as extraction models. We rank five responses from five models using GPT-4o, Llama3.1-70B, and Llama3.3-70B as our judge models. The the correlation between different judge models and methods are included in Table~\ref{tab:correlation_results}.
The judge prompts follow a fixed template with seven placeholders, where the ontology placeholder represents entity and relation types. Formally, let the input set for the LLM judge prompts be \(\{O, C, R_{m_1},R_{m_2},R_{m_3},R_{m_4}, R_{m_5} \}\), where \(O\) is the ontology set, \(C\) is the set of test contexts, and 
$ R_{m_1}, \cdots, R_{m_5}$ are the five sets of candidate answers from the five extractor models. The judge LLM produces a verdict (output as a ranking):
\begin{equation}
V = \mathrm{LLM}\bigl(\{O, C, R_{m_1},R_{m_2},R_{m_3},R_{m_4}, R_{m_5}\}\bigr),
\end{equation}
assigning a rank from best (1) to worst (5) based on predefined instructions and ranking criteria.

To mitigate evaluation biases, we design three judge prompt templates: (1) Basic Judge Prompt, (2) Fair Judge Prompt, and (3) Randomized Fair Judge Prompt. These templates differ in their instructions and ranking methodologies. Fair Judge Prompt enforces explicit reasoning criteria to mitigate position bias, a known issue when LLMs evaluate multiple candidate answers simultaneously \citep{li2024split,shi2024judging}. Randomized Fair Judge Prompt further reduces this bias by randomizing the position of candidate answers, ensuring that response order does not influence rankings. 

Once the optimal judge LLM setting is determined, we adopt it as the reference-free evaluator to identify the best answer from each extraction, leveraging its reasoning process. Detailed prompt templates and an example of the reasoning process used for evaluation are provided in Appendix.

\section{Experimental Results}
\label{sec:Results}
We assess the effectiveness of our knowledge triple extraction method through \textit{reference-based} and \textit{reference-free} evaluations. Reference-based evaluation compares extracted triples against our curated dataset, while reference-free evaluation relies on LLM judges to assess generated triples without reference data. A key aspect of our analysis is to examine the correlation between these two evaluation paradigms to evaluate the reliability of LLM-based judgments. In addition, we investigate how different prompt strategies influence extraction performance between models.

\begin{figure}[h]
    \centering
    \includegraphics[width=1.0\columnwidth]{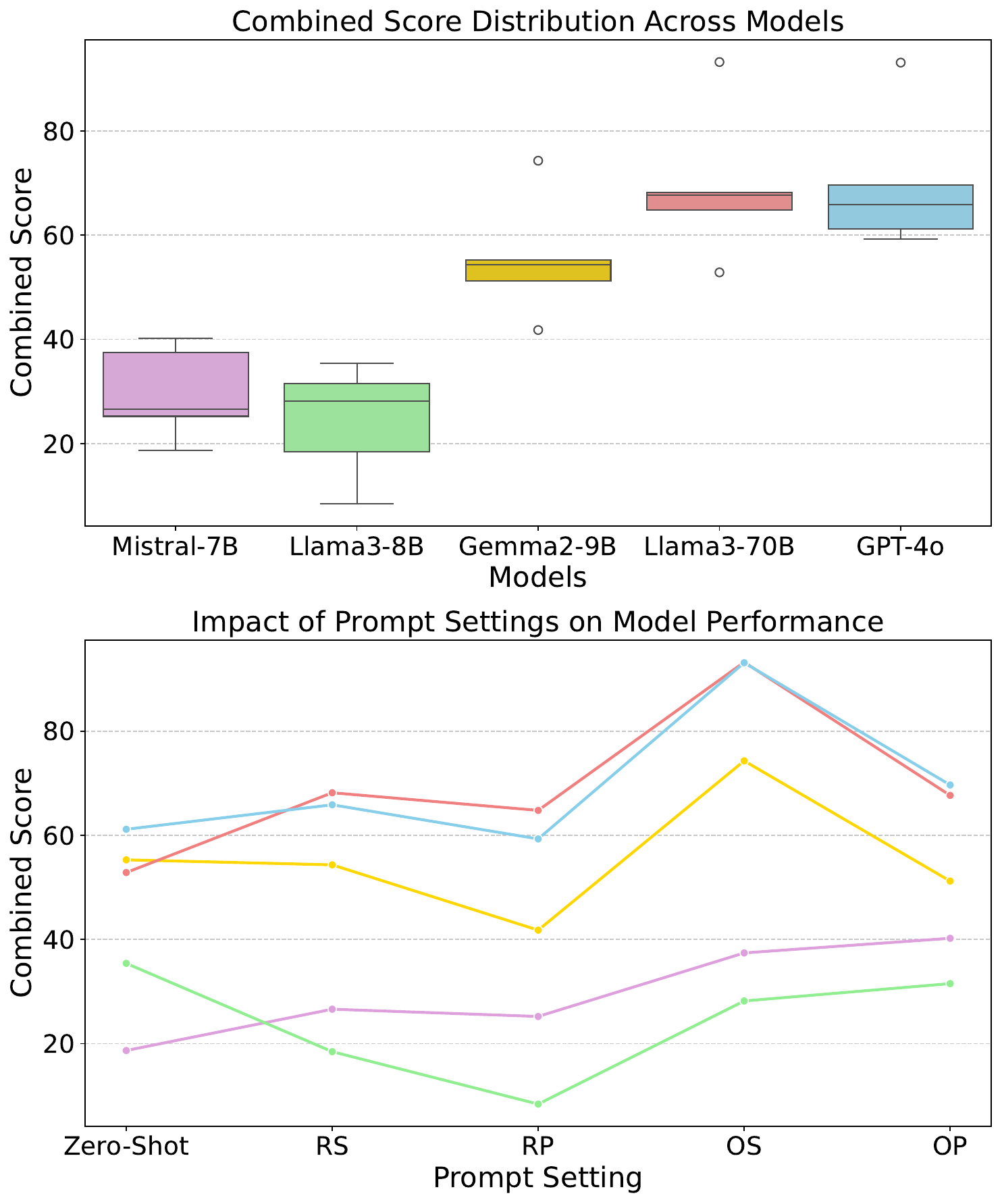}
    \caption{The combined visualization illustrates the impact of model selection and prompt strategy on extraction performance. Abbreviations on x-axis are about One-shot with: RS = Random Sentences; RP = Random Paragraphs; OS =  Ontology-Aligned Sentence; OP = Ontology-Aligned Paragraph.
    The top Combined Score is achieved by Llama3-70B (93.24) closely followed by GPT-4o (93.13), both with OS prompt setting.}
    \label{fig:Combined}
    \rmspace
\end{figure}

\paragraph{Reference-Based Evaluation} We employ five LLMs as extractor models: Llama3-70B, GPT-4o, Gemma-9B, Llama-8B and Mistral-7B.
Figure~\ref{fig:Combined} illustrates the impact of model selection and prompt strategy on extraction performance. The box plot (top) shows the distribution of \textit{Combined Scores} across models. Llama3-70B achieves the highest overall performance score, closely followed by GPT-4o. Gemma2-9B demonstrates moderate performance, while Llama3-8B and Mistral-7B receive lowest overall scores. The line plot (bottom) highlights prompt effects, with OS and OP yielding the highest scores across most models, supporting the effectiveness of ontology-aligned prompting. These findings reinforce our hypothesis that ontology-aligned prompts enhance extraction accuracy.

\begin{figure*}[h]
    \centering
    \includegraphics[width=\textwidth]{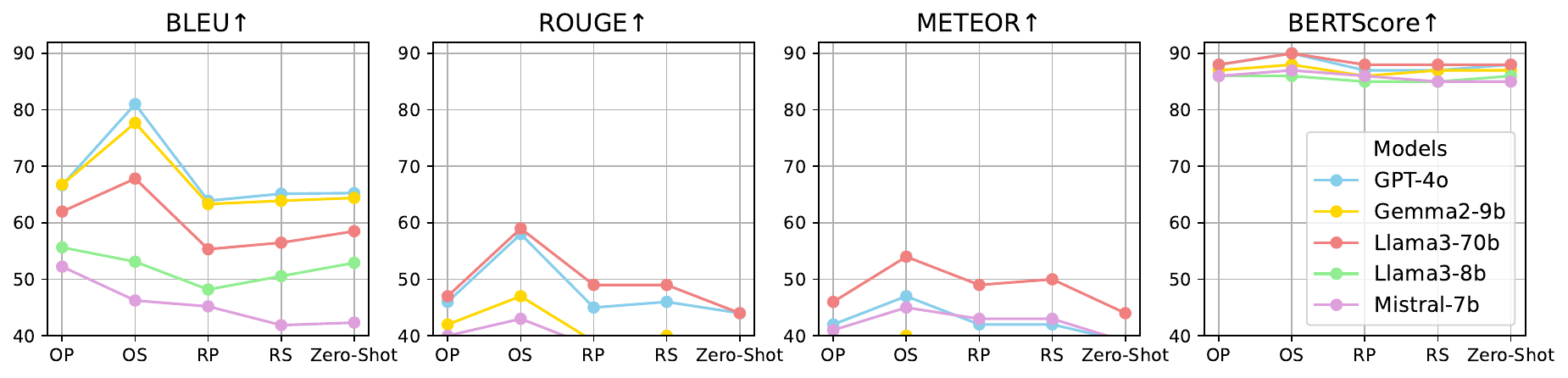}
    \caption{Accuracy metrics scores across prompt types for each model. OS demonstration prompts consistently result in the best accuracy across all four metrics and all five models. \textit{Note: ROUGE, METEOR are scaled by 150, BERTScore by 100 for better visibility.}}
    \label{fig:Accuracy}
    \rmspace
\end{figure*}

Figure~\ref{fig:Accuracy} further breaks down the performance of five models across four accuracy evaluation metrics: BLEU, ROUGE, METEOR, and BERTScore. OS demonstration prompts consistently result in the best accuracy across all four metrics and all five models, highlighting their effectiveness for the triple extraction task. BLEU scores peak with OS prompts, with GPT-4o achieving the best performance. ROUGE results show a similar trend, with GPT-4o and Llama3-70B excelling in the OS prompt setting. METEOR follows the same pattern as BLEU and ROUGE, reinforcing the advantages of OS prompts. BERTScore, which measures semantic similarity, shows high clustering across models, suggesting minimal differentiation in performance.
Overall, OS demonstration prompts consistently enhance extraction accuracy across all models and metrics. 

\begin{figure*}[h]
    \centering
    \includegraphics[width=\textwidth]{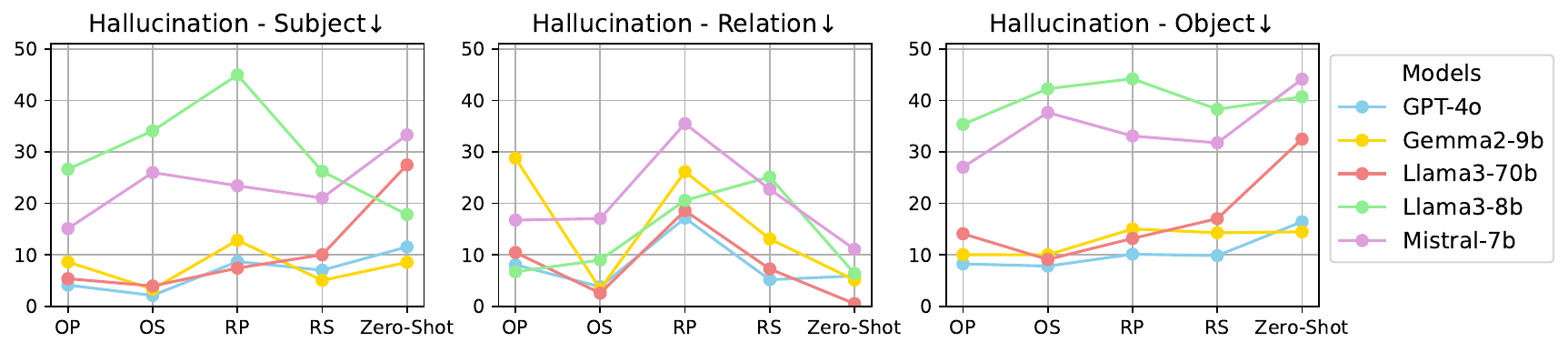}
    \caption{Hallucination rate of subject, relation, and object across prompt types for each model.}
    \label{fig:hallucination}
    \rmspace
\end{figure*}

Figure~\ref{fig:hallucination} presents the hallucination rates for subjects, relations, and objects across different models and prompt settings. GPT-4o, Llama3-70B, and Gemma2-9B exhibit lower hallucination rates for subjects and objects, while Llama3-8B and Mistral-7B tend to have higher hallucination rates. For high-performing models, the OS prompt type generally helps reduce hallucination, whereas RP prompts tend to increase it. Zero-Shot prompts often lead to increased hallucination for subjects and objects across models. Interestingly, however, Zero-Shot prompts show lower hallucination rates for relations, which may be due to the additional demonstrations in other prompts introducing noise that negatively impacts relation extraction. 
Most models exhibit high format conformance across all prompt types, with only Gemma-9B under the Zero-Shot prompt scoring below 80\%. GPT-4o and Llama3-70B consistently achieve FC above 95\% across all prompt types, demonstrating superior adherence to the expected format. 

\noindent \textbf{Reference-Free Evaluation}
Results show that the OS prompt consistently achieves the highest performance across most models, so for the reference-free ranking experiments we only consider five models under OS prompt setting. To assess the alignment between reference-based and reference-free rankings, we compute correlations between the rankings derived from the \textit{Combined Score} calculated based on references, and the rankings derived from the \textit{Expectation Score} based on LLM judges. 

\paragraph{Expectation Score} As each extractor model received multiple rankings from different judge models, we compute a single  \textit{Expectation Score} per extractor model $m$. This score is defined as:
\begin{equation}
    \textit{Expectation Score } E(m) = \frac{\sum_{i=1}^{\mu} \big(i \times P_m(i)\big)}{\sum_{i=1}^{\mu} P_m(i)}
\end{equation}

where \( i \) represents a specific rank, \( P_m(i) \) denotes the number of times the model $m$ was assigned rank \( i \). In our case from $i=1,\cdots, \mu$ and $\mu=5$ for the five extractor models. $E$ provides a weighted average that reflects the overall tendency of judge models to place an extractor model at a particular rank. The extractor models are ranked such that the one with lowest \textit{Expectation Score} is ranked the highest and vice versa. To systematically analyze the consistency and reliability of the judge methods, we compute the \textit{Expectation Scores} of extractor models separately for Basic Judge, Fair Judge, and Randomized Fair Judge. 

\paragraph{Correlation Between LLM Judged and Reference-Based Rankings}

To assess the reliability of LLM judges, we compute Spearman’s correlation (\(\rho\)) and Kendall’s Tau (\(\tau\)) to quantify the alignment between LLMs judged rankings and reference-based rankings. \(\rho\) measures the monotonic relationship between rankings, where values close to 1 indicate strong agreement. \(\tau\) evaluates ranking concordance by analyzing the number of concordant and discordant rank pairs, making it particularly useful for detecting minor positional changes. The results of our iterative ranking experiments, shown in Table~\ref{tab:correlation_results}, demonstrate how different judge methods impact ranking alignment.

Our findings reveal that introducing randomization significantly enhances ranking consistency for GPT-4o, improving from \(\rho = 0.4\) (Basic) to \(\rho = 1.0\) (Randomized). In contrast, Llama3.1-70B shows no improvement across judge methods, indicating persistent positional bias. Llama3.3-70B exhibits weaker alignment overall, with minor improvements under Fair and Randomized judging. These results suggest that while randomization effectively mitigates positional bias for GPT-4o, its impact varies across models.

\begin{table}
\caption{The correlation results for different judge models, judge methods.}
\label{tab:correlation_results}
\centering
\resizebox{\linewidth}{!}{%
\begin{tabular}{@{}lccc@{}}
\toprule
\textbf{Judge Model} & \textbf{Judge Method} & \textbf{Spearman’s Correlation} & \textbf{Kendall’s Tau} \\ 
\midrule
\multirow{3}{*}{GPT-4o} & Basic       & 0.4 & 0.4 \\ 
                        & \cellcolor[HTML]{CBCEFB}Fair  & \cellcolor[HTML]{CBCEFB}0.9 & \cellcolor[HTML]{CBCEFB}0.8 \\ 
                        & \cellcolor[HTML]{9698ED}Randomized  & \cellcolor[HTML]{9698ED}1.0 & \cellcolor[HTML]{9698ED}1.0 \\ 
\midrule
\multirow{3}{*}{Llama3.1-70B} & Basic       & 0.4 & 0.4 \\ 
                               & Fair        & 0.4 & 0.4 \\ 
                               & Randomized  & 0.4 & 0.4 \\ 
\midrule
\multirow{3}{*}{Llama3.3-70B} & Basic       & 0.0 & -0.2 \\ 
                               & Fair        & 0.3 & 0.2 \\ 
                               & Randomized  & 0.2 & 0.2 \\ 
\bottomrule
\end{tabular}%
}
\rmspace
\end{table}

The ranking experiments identify GPT-4o with Randomized Fair Judge as the optimal judge LLM setting for our evaluation task. We further apply this setting to identify the optimal triples from each extraction output using GPT-4o with the Randomized Fair Judge method, aggregating the top-ranked triples across all extractions to generate the final output. We compare these aggregated results against the test dataset, yielding a \textit{Combined Score} of 83.93. This achieves 90\% of the current best score 93.24 (see figure \ref{fig:Combined}), reinforcing the effectiveness of our reference-free LLM-as-a-Judge paradigm as a viable alternative to conventional reference-based evaluation methods. Future research could refine this approach by exploring additional LLMs and judge strategies to approximate a even better judge LLM setting for knowledge triple extraction tasks.

\begin{figure*}[h]
    \centering
    \includegraphics[width=\linewidth]{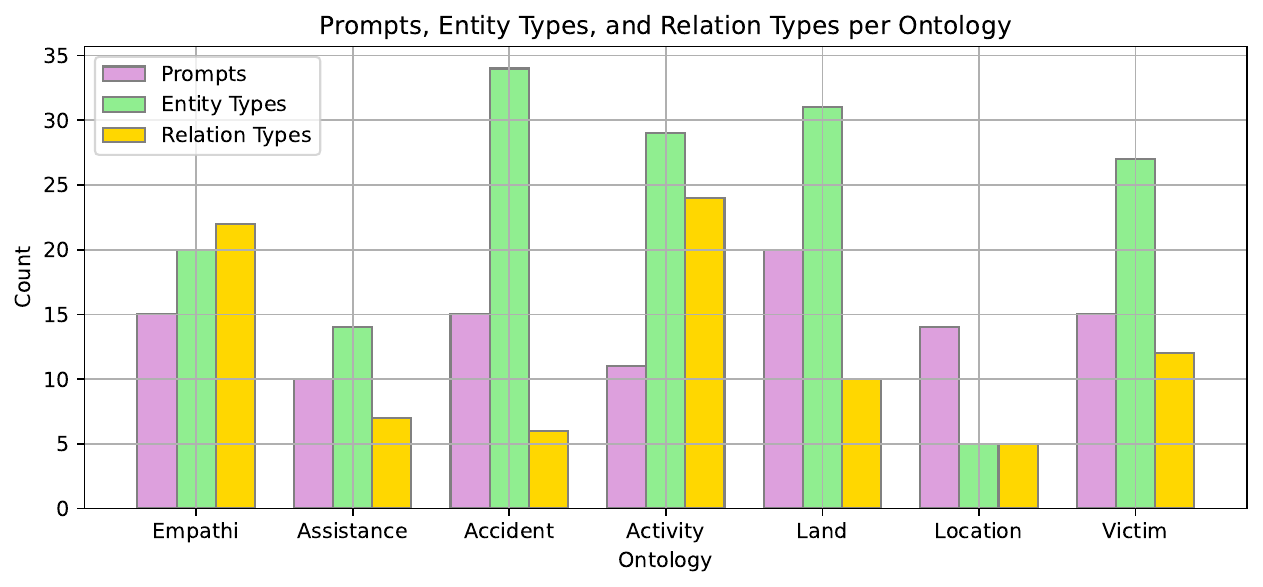}
    \caption{Number of prompts, entity types, and relation types per the utiized ontologies from IMSMA Core and Empathi.}
    \label{fig:numprompts}
    
\end{figure*}
\subsection{Additional Information and Insights for the HMA Ontology}
\label{app:ontology}

We visualize the statistics regarding the combined ontologies. Fig.~\ref{fig:numprompts} includes the number of prompts, entity types, and relation types per ontology. 

The Empathi~\cite{8665539} ontology is designed for ``Emergency Managing and Planning about Hazard Crises''. It is the most advanced ontology that is openly-available in the research domain for hazard crises, as an outcome of a research study. However, this ontology is designed for the broader concept of disasters as opposed to being specialized in the HMA domain. For instance, it has many concepts such as hurricanes or firefighters that are outside of the scope of the demining-relevant disasters such as mine explosions. The concepts (entity types and relation types) that are relevant to the demining are manually filtered and used as the ontology. Thus, the number of entities and relations are after the filtering process. GICHD's new data schema IMSMA Core (previous standard was called IMSMA-NG~\cite{imsmang}, where NG stands for ``Next Generation'') is the agreed standard between mine agencies of governments and non-governmental organizations (NGOs). IMSMA Core serves the data model for databases of mine agencies globally. On the other hand, the number of entities and relation types are limited.  

We consider the separate portions of the GICHD data model, which are all related to the mine action but logically separated as \textit{``Assistance, Accident, Activity, Land, Location, and Victim''} ontologies. We utilize both as the best state-of-the-art models and convert them into relatively small ontologies. On the other hand, we utilize the concepts from all of the 7 ontologies in our prompt generation process for LLMs, resulting in different number of prompts per ontology. The proposed ``HMA Ontology'' can be extended with more special entity and relation types based on special needs of the organizations.
\balance

\section{Conclusion}
\label{sec:Conclusion}
TextMineX addresses the need for knowledge extraction from HMA reports by LLMs and domain ontologies to transform unstructured technical reports into structured knowledge triples. This paper introduces the curated data of HMA reports, ontology, and LLM pipeline, aiming to enable innovations for mine action and natural language processing.

\clearpage

\section*{Limitations}

While TextMineX provides the first dataset, evaluation framework and pipeline that demonstrates the feasibility of an LLM-driven extraction pipeline in the specialized domain of humanitarian demining, several limitations are considered to be addressed in the future, including further extension of the dataset. Our evaluation relies on 100 prompt–response examples (yielding 1,095 unique triples). 

Assembling and annotating demining data demand extensive domain expertise, so even this modest set offers valuable proof-of-concept insights, but we plan to extend to larger, multilingual collections in future work. Second, this work focuses on reports that are written only in English. In the future, we plan to extend the work to include other languages, because mine-action data are often produced in multiple languages, such as Arabic, French or Ukrainian, depending on the conflict region. Making TextMineX multilingual is, therefore, needed to ensure more inclusiveness. 

In this paper, we focus on humanitarian demining and actively collaborate with demining agencies. Thus, the scope of the evaluation framework and pipeline is limited to the demining domain, whereas the applicability of the pipeline in similar domains (e.g, natural disaster management) could be tested as a potential future work.

\section*{Ethical Considerations}
Humanitarian demining is a high importance decision making domain where incorrect or hallucinated triples can misinform planning and lead to wasted time and resources. To address this, TextMineX combines reference based validation with a bias aware LLM as Judge framework and publishes all prompts and decoding settings for full transparency. We also engage landmine clearance experts throughout development to review and validate outputs. TextMineX is not used to locate mines, as safety remains governed by established GICHD standards\footnote{GICHD Mine Action Standards, \url{https://www.gichd.org/our-response/mine-action-standards/}}, and instead supports expert analysis and planning. By documenting our methods and keeping an expert in the loop, we aim to minimize misinformation and ensure responsible AI deployment in demining operations.  

The dataset used in this study consists of publicly available or institutionally provided humanitarian demining reports. These reports were reviewed to ensure they do not contain personally identifiable information (PII) or offensive content. Our usage of the data adheres to privacy standards and is strictly confined to research contexts.

All datasets used in this study were accessed under conditions permitting research use. The curated demining report dataset we constructed is intended solely for academic and research purposes and complies with the original access and licensing conditions. The ontology and pipeline components developed in TextMineX are likewise designed for research and evaluation within humanitarian domains. We do not support or promote deployment of these artifacts in operational or commercial contexts without further validation and ethical review.

AI usage: AI assistance tool is utilized for grammar/spelling check of the paper.

\bibliography{bibliography}

\newpage
\onecolumn
\appendix
\label{Appendix}
\section{Detailed Discussion of Related Work}
\label{app:rel-work}

\subsection{Automated Knowledge Graph Construction}
Constructing knowledge graphs (KGs) from text is a well-established problem \cite{weikum2021machine}. The methods for constructing KGs are either schema-free \cite{pei2023abstractive,kotnis-etal-2022-milie,gashteovski2019opiec} or are fixed either to general-purpose schemas \cite{zhang2024extract,josifoski-etal-2022-genie,broscheit2017openie} or to domain-specific schemas \cite{dagdelen2024structured,lauscher2019minscie,broscheit2017openie}. Both the schema-free methods and the methods that are tied to general-purpose schemas do not address the complexities of the domain at hand, which is why we worked on a specific data, benchmarks and methods for the HMA domain.

\subsection{LLMs for Knowledge Graph Construction}
Prior work has explored the use of LLMs for constructing KGs from text \cite{pan2023large}. One line of work focuses on harvesting knowledge from pretrained LLMs without the use of annotated data via zero-shot iterative prompting \cite{hu2025enabling,carta2023iterative,hao2022bertnet} or in-context learning (ICL) \cite{brown2020language,min2022rethinking,liu2023tabular}. However, these methods are not well-suited for the construction of domain-specific knowledge graphs. 

For these reasons, there have been methods for constructing domain-specific KGs by the use of LLMs \cite{chen2024sac}. For example, there are LLM methods for constructing KGs in various domains, such as medicine \cite{arsenyan2023large}, finance \cite{li2025finkario}, cyber-thread intelligence \cite{huang2024ctikg} or climate science \cite{pan2025taxonomy}. To the best of our knowledge, however, there is no prior work on KG construction for the humanitarian mine action domain.

\subsection{Ontology-Guided for Knowledge Graph Construction}

Ontology-guided (a.k.a.~schema-driven) methods have been shown to be effective, though they are often limited to single-sentence input and, consequently, overlook context-level reasoning \cite{cauter2024ontology}. Text2KGBench \cite{mihindukulasooriya2023text2kgbench} is a benchmark and evaluation framework for extracting triples from text, which are bound to a predefined ontology. However, it relies on toy data schemas, which limits their applicability to the complex, domain-specific HMA documents. 

Other line of work focuses on reasoning and validation within already structured knowledge graphs, not extraction from free-form text as with TextMineX \cite{liu2025ontology,regino2025can}. Their ontological constraints apply post-hoc to existing triples, whereas TextMineX uses the ontology during extraction to guide triple formation and reduce hallucinations.

Finally, none of these works address bias-aware evaluation: they rely on deterministic metrics or small-scale manual validation. TextMineX introduces a bias-aware LLM-as-Judge evaluation framework that explicitly mitigates positional bias in reference-free scoring—an issue absent from prior ontology-guided pipelines.

\subsection{Multi-Faceted Knowledge Graph Evaluation}

They are typically evaluated with a single score metric, such as accuracy \cite{eisenstein2019introduction}. In recent years, however, the NLP community has increasingly been pointing out that such evaluation approach can obscure the model performance nuances, thus revealing misleading results \cite{ribeiro2020beyond}. 

To mitigate these effects, researchers have turned to multiple metric evaluations, which reveal more complete picture about the performance of the models \cite{vickers2024we,opitz2024closer,jain2023multi,deutsch2022limitations,peyrard2021better,liu2021explainaboard}. Particularly within the field of information extraction, such multi-faceted evaluations are already well established \cite{radevski2023linking,friedrich-etal-2022-annie,gashteovski-etal-2022-benchie,fu2020interpretable}. Likewise, KG-related work also employs such multi-faceted evaluations, such as for the task of link prediction \cite{gastinger2023comparing,widjaja2022kgxboard,rim2021behavioral,meilicke2018fine}.

In order to evaluate aspects of the models that are important to users or that need immediate human verifications, another line of work have used manual evaluations \cite{bojic2023hierarchical,novikova2018rankme}. This could be done with user studies that target the specific aspects that one tries to measure \cite{radevski2025synthesizing,schuff2023human,kotnis2022human} or by simply manually validating the final results \cite{Thomson2024,Lee2021HumanEO}. Such manual evaluations have been practiced for both information extraction research works \cite{Sainz2023GoLLIEAG,Gashteovski2020OnAO,Bhardwaj2019CaRBAC}, as well as for KG prediction tasks, such as link prediction \cite{Carriero2024HumanEO,Xu2024AHE,Zhou2022RethinkingKG}. Manual evaluations, however, are both expensive and not scalable, thus making them practically infeasible.

To avoid the pitfalls of the narrow single-score metrics, as well as the scalability issues of manual evaluations, we propose a multi-faceted evaluation framework that incorporates several aspects of the information extraction problem, fused into the final score.

\section{In-Context Learning Prompt Example}

The below is an example of one-shot prompt with the RS prompt setting. All in-context learning prompts are stored in CSV files as part of the supplementary material for easier reproducibility.

\begin{tcolorbox}[colback=brown!10, colframe=brown!80, title=One-shot with Random Sentence Demonstration (RS), width=\linewidth, boxrule=0.5mm, arc=3mm, auto outer arc]
\small
\textbf{Instruction:} 

Extract and list only the triples from the following sentence based on the specified entity types and relation types. Do not include any explanatory or intermediate text in your output. 
In the output, only include the triples in the given output format: relation(subject, object). Attempt to extract as many entities and relations as you can. 

\textbf{Entity Types:} 

AdministrativeArea, Association, Location, Organisation, MedicalFacility

\textbf{Relation Types:} 

hasAdministrativeArea, hasAssociation, hasLocation, hasOrganisation, locatedNear

\textbf{Example:}

\textbf{Sentence:} 

The accidental detonation of old wartime munitions causes significant infrastructure damage to the nearby village roads and buildings.

\textbf{Output:} 

CausedBy(infrastructure damage, old wartime munitions)

\textbf{Context:}
On Thursday, March 16, 2023, at CMAC Headquarters in Phnom Penh, 
Delegate of the Royal Government in charge as Director General of CMAC, met with a 
delegation from the Japan International Cooperation Agency (JICA) 
General Director of Governance and Peacebuilding Department. During the meeting, the JICA 
side briefed on the results of its cooperation with CMAC, in particular training for Ukraine with 
good results. 
\end{tcolorbox}

\section{LLM Judge Prompts}

\begin{tcolorbox}[colback=brown!10, colframe=brown!80, title=Basic Judge Prompt, width=\linewidth, boxrule=0.5mm, arc=3mm, auto outer arc]
\small
\textbf{Instruction:} 

You are a judge who ranks five models from 1 to 5 on a triple extraction task. You must assign 1 to the model with the best answer and 5 to the model with the worst answer. Your ranking should be provided directly in this format: [1: model x; 2: model x; 3: model x; 4: model x; 5: model x].

\textbf{Ranking Criteria:}

\textbf{Correctness:} 

The triples must conform to the format relation(subject, object) and must accurately reflect relationships stated in the context. Models with significant formatting errors should be penalized.

\textbf{Coverage:} 

The number of correct triples extracted. More accurate triples are better, but avoid penalizing slight redundancies unless they detract from the overall relevance.

\textbf{Relevance:} 

The triples must be relevant to the specified entity and relation types and should align well with the specific context provided.

\textbf{Edge Cases:} 

If a model extracts many triples but includes incorrect or redundant ones, balance accuracy and redundancy in your ranking. Correctness should be prioritized, followed by Relevance, then Coverage.

\textbf{Entity Types:} \{entity\_types\}

\textbf{Relation Types:} \{relation\_types\}

\textbf{Context:} \{Context\}

\textbf{Model Outputs:} \{model 1 output\} \{model 2 output\} \{model 3 output\} \{model 4 output\} \{model 5 output\}

\textbf{Your ranking:}
\end{tcolorbox}

The below are two example prompts used during the experimental study: 1) Basic judge prompt, and 2) (Randomized) Fair Judge Prompt. These methods are explained in Sec.~\ref{sec:Methodology}. The latter prompt example below includes both cases of regular and randomized fair judge prompts at once. The only difference is the shuffling of the positions of candidate answers in ``Model Outputs'' part of the prompt. All LLM judge prompts are stored in CSV files as part of the supplementary material. The prompts can be used for reproducing as well as applying in different datasets (without additional annotation efforts).

\begin{tcolorbox}[colback=brown!10, colframe=brown!80, title= (Randomized) Fair Judge Prompt, width=\linewidth, boxrule=0.5mm, arc=3mm, auto outer arc]
\small
\textbf{Instruction:} 

You are a judge tasked with evaluating and ranking five models based on their performance in a \textbf{triple extraction task}. Your role is to ensure \textbf{fairness, impartiality, and accuracy} by independently evaluating each model's output without any positional bias. Do not assume that the first model is better or worse simply because of its position—all models must be treated equally.

\textbf{Evaluation Guidelines:}
\begin{enumerate}[leftmargin=0cm]
    \item \textbf{Independence of Evaluation}:
    
    Evaluate each model \textbf{independently} without comparing it to others until all models are scored. Avoid assumptions based on position or order in the list.

    \item \textbf{Evaluation Criteria:}
    \begin{enumerate}[leftmargin=0cm]
        \item \textbf{Correctness of Triples (Highest Priority):}
        \begin{itemize}[leftmargin=0cm]
            \item Triples must strictly conform to the format \texttt{relation(subject, object)}.
            \item Relationships must match the \textbf{Given Relation Types} provided below.
            \item Triples containing fabricated or hallucinated relationships must result in a significant penalty.
        \end{itemize}
        \item \textbf{Relevance:}
        \begin{itemize}[leftmargin=0cm]
            \item Triples must accurately reflect relationships mentioned in the \textbf{Context}.
            \item Irrelevant triples or hallucinations must receive a lower score.
        \end{itemize}
        \item \textbf{Coverage:}
        \begin{itemize}[leftmargin=0cm]
            \item The number of correct triples extracted. Higher coverage is better \textbf{only} if triples meet correctness and relevance criteria.
        \end{itemize}
    \end{enumerate}

    \item \textbf{Ranking Process:}
    \begin{itemize}[leftmargin=0cm]
        \item \textbf{Step 1:} Independently evaluate each model's output and assign scores (from 1 to 10) for each criterion: Correctness, Relevance, and Coverage. Summarize the total score for each model.
        \item \textbf{Step 2:} Rank all five models from 1 (best) to 5 (worst) based solely on their total scores.
        \item Break ties by prioritizing \textbf{Correctness} first, then \textbf{Relevance}, and finally \textbf{Coverage}.
    \end{itemize}
\end{enumerate}

\textbf{Important Note:} 

Treat all models equally regardless of position. Your goal is to be an impartial judge. Avoid positional bias and ensure the evaluation reflects only the quality of the model outputs.

\textbf{Output Format:} 

Provide your output strictly in this format:\\
Evaluation: \\
Model 1: Correctness = X, Relevance = X, Coverage = X, Total = Y \\
Model 2: Correctness = X, Relevance = X, Coverage = X, Total = Y \\
Model 3: Correctness = X, Relevance = X, Coverage = X, Total = Y \\
Model 4: Correctness = X, Relevance = X, Coverage = X, Total = Y \\
Model 5: Correctness = X, Relevance = X, Coverage = X, Total = Y \\
Ranking: 
[1: Model X; 2: Model X; 3: Model X; 4: Model X; 5: Model X]

\textbf{Entity Types:} \{entity\_types\}

\textbf{Relation Types:} \{relation\_types\}

\textbf{Context:} \{Context\}

\textbf{Model Outputs:} \{model 1 output\} \{model 2 output\} \{model 3 output\} \{model 4 output\} \{model 5 output\}\\
\textcolor{red}{(For Randomized Fair Judge Prompts, the positions of candidate answers in Model Outputs are randomly shuffled.)}\\
\textbf{Your evaluation and ranking:}
\end{tcolorbox}

\section{Evaluation Reasoning Process}

\begin{tcolorbox}[colback=brown!10, colframe=brown!80, title=Output of GPT-4o with Randomized Judge Prompt for Extraction Prompt 89, width=\linewidth, boxrule=0.5mm, arc=3mm, auto outer arc]
\small
\subsection*{Model's Answer for Prompt 89}

\subsection*{Evaluation}

\subsubsection*{Model 1}
\begin{itemize}
    \item \textbf{Correctness} = 4
    \begin{itemize}
        \item Many relations are either incorrect or irrelevant to the context (e.g., Monitoring, Advises, Coordinates), leading to significant penalties.
    \end{itemize}
    \item \textbf{Relevance} = 5
    \begin{itemize}
        \item The model produces correct triples but includes several irrelevant triples not related to the context (e.g., PhysicalHealth).
    \end{itemize}
    \item \textbf{Coverage} = 7
    \begin{itemize}
        \item High number of triples extracted, but this value is diminished by the correctness and relevance issues.
    \end{itemize}
    \item \textbf{Total} = 4 + 5 + 7 = 16
\end{itemize}

\subsubsection*{Model 2}
\begin{itemize}
    \item \textbf{Correctness} = 6
    \begin{itemize}
        \item Includes both relevant and some fabricated or unclear relations (e.g., Triggers(InfrastructureDamage, old wartime munitions)).
    \end{itemize}
    \item \textbf{Relevance} = 6
    \begin{itemize}
        \item Many triples are relevant, though some irrelevant or vague triples are present.
    \end{itemize}
    \item \textbf{Coverage} = 7
    \begin{itemize}
        \item Covers a broad range of the context accurately but is diminished slightly by prioritizing the correctness and relevance.
    \end{itemize}
    \item \textbf{Total} = 6 + 6 + 7 = 19
\end{itemize}

\subsubsection*{Model 3}
\begin{itemize}
    \item \textbf{Correctness} = 7
    \begin{itemize}
        \item Most triples are accurate and correctly formed, with fewer fabricated relations.
    \end{itemize}
    \item \textbf{Relevance} = 7
    \begin{itemize}
        \item Majority of the triples are highly relevant to the context. Some triples are slightly incoherent but mostly pertinent.
    \end{itemize}
    \item \textbf{Coverage} = 6
    \begin{itemize}
        \item Good coverage of the context, though a few needed extractions were missing.
    \end{itemize}
    \item \textbf{Total} = 7 + 7 + 6 = 20
\end{itemize}

\subsubsection*{Model 4}
\begin{itemize}
    \item \textbf{Correctness} = 6
    \begin{itemize}
        \item Contains some inaccuracies but generally provides correctly formulated triples.
    \end{itemize}
    \item \textbf{Relevance} = 6
    \begin{itemize}
        \item Various triples are relevant to the context, but some are less directly related.
    \end{itemize}
    \item \textbf{Coverage} = 5
    \begin{itemize}
        \item Moderate coverage but some key elements are missed or less comprehensive.
    \end{itemize}
    \item \textbf{Total} = 6 + 6 + 5 = 17
\end{itemize}

\subsubsection*{Model 5}
\begin{itemize}
    \item \textbf{Correctness} = 8
    \begin{itemize}
        \item High accuracy in triple formation with minimal fabrications.
    \end{itemize}
    \item \textbf{Relevance} = 8
    \begin{itemize}
        \item Highly relevant triples, capturing the context well without introducing unrelated information.
    \end{itemize}
    \item \textbf{Coverage} = 8
    \begin{itemize}
        \item Extensive and accurate coverage of the context, handling the core relationships well.
    \end{itemize}
    \item \textbf{Total} = 8 + 8 + 8 = 24
\end{itemize}

\subsection*{Ranking}
\begin{enumerate}
    \item \textbf{Model 5}
    \item \textbf{Model 3}
    \item \textbf{Model 2}
    \item \textbf{Model 4}
    \item \textbf{Model 1}
\end{enumerate}

\end{tcolorbox}

\clearpage

\end{document}